\documentclass[10pt,twocolumn,letterpaper]{article}

\usepackage[pagenumbers]{cvpr} 

\usepackage{graphicx}
\usepackage{amsmath}
\usepackage{amssymb}
\usepackage[title]{appendix}
\usepackage{xspace}
\usepackage{mathtools}
\usepackage{bbding}
\usepackage{amsmath}
\usepackage{amsthm}
\usepackage{booktabs}
\usepackage{verbatim}
\usepackage{multirow}
\usepackage{makecell}
\usepackage{algorithm}
\usepackage{algorithmic}
\usepackage{times}
\usepackage{epsfig}
\usepackage{pifont}
\usepackage{booktabs}
\usepackage{capt-of}
\usepackage{colortbl}
\usepackage{dsfont}
\usepackage{comment}
\usepackage[dvipsnames]{xcolor}
\usepackage{pifont}
\usepackage{bbm}
\usepackage{url}
\usepackage{nicefrac}
\usepackage{colortbl}
\usepackage{cases}
\usepackage{graphicx}
\usepackage{amsmath}
\usepackage{amssymb}
\usepackage{booktabs}
\usepackage{graphicx} 
\usepackage{booktabs}
\usepackage{multirow}
\usepackage[title]{appendix}
\usepackage{xspace}
\usepackage{mathtools}
\usepackage{bbding}

\usepackage{graphicx}
\usepackage{amsmath}
\usepackage{amsthm}
\usepackage{booktabs}
\usepackage{algorithm}
\usepackage{algorithmic}
\usepackage{verbatim}
\usepackage{multirow}
\usepackage{makecell}
\usepackage{epsfig}
\usepackage{graphicx}
\usepackage{amsmath}
\usepackage{amssymb}
\usepackage{bbm}
\usepackage{pifont}
\usepackage{booktabs}
\usepackage{multirow}
\usepackage{capt-of}
\usepackage{makecell}
\usepackage{colortbl}
\usepackage{dsfont}
\usepackage{comment}
\usepackage[dvipsnames]{xcolor}
\usepackage{pifont}
\usepackage{bbm}
\usepackage{url}
\usepackage{nicefrac}

\usepackage{colortbl}
\definecolor{mygray}{gray}{.9}

\definecolor{cvprblue}{rgb}{0.21,0.49,0.74}
\usepackage[pagebackref,breaklinks,colorlinks,citecolor=cvprblue]{hyperref}

\usepackage[capitalize]{cleveref}
\crefname{section}{Sec.}{Secs.}
\Crefname{section}{Section}{Sections}
\Crefname{table}{Table}{Tables}
\crefname{table}{Tab.}{Tabs.}

\definecolor{baselinecolor}{gray}{.9}
\definecolor{darkgreen}{rgb}{0.13, 0.55, 0.13}

\newcommand{\tablestyle}[2]{\setlength{\tabcolsep}{#1}\renewcommand{\arraystretch}{#2}\centering\footnotesize}

\renewcommand{\paragraph}[1]{\vspace{1.25mm}\noindent\textbf{#1}}

\let\originalleft\left
\let\originalright\right
\renewcommand{\left}{\mathopen{}\mathclose\bgroup\originalleft}
\renewcommand{\right}{\aftergroup\egroup\originalright}

\usepackage{caption}
\begin{document}


\title{DreamMask: Boosting Open-vocabulary Panoptic Segmentation \\ 
with Synthetic Data}

\author{
    Yuanpeng Tu$^{1}$ \quad
    Xi Chen$^{1}$ \quad
    Ser-Nam Lim$^{2}$ \quad
    Hengshuang Zhao$^{1,\dagger}$\\[2pt]
    $^{1}$HKU \quad
    $^{2}$UCF \quad
    \\
    \textit{\href{https://yuanpengtu.github.io/Dreammask-Page/}{https://yuanpengtu.github.io/Dreammask-Page/}}
}

\maketitle

\let\thefootnote\relax\footnotetext{$\dagger$ Corresponding author.}

\begin{abstract}

Open-vocabulary panoptic segmentation has received significant attention due to its applicability in the real world.
Despite claims of robust generalization, we find that the advancements of previous works are attributed mainly to trained categories, exposing a lack of generalization to novel classes.
In this paper, we explore boosting existing models from a data-centric perspective. 
We propose DreamMask, which systematically explores how to generate training data in the open-vocabulary setting, and how to train the model with both real and synthetic data. 
For the first part, we propose an automatic data generation pipeline with off-the-shelf models. We propose crucial designs for vocabulary expansion, layout arrangement, data filtering, etc.  
Equipped with these techniques, our generated data could significantly outperform the manually collected web data. 
To train the model with generated data, a synthetic-real alignment loss is designed to bridge the representation gap, bringing noticeable improvements across multiple benchmarks.
In general, DreamMask significantly simplifies the collection of large-scale training data, serving as a plug-and-play enhancement for existing methods. For instance, when trained on COCO and tested on ADE20K, the model equipped with DreamMask outperforms the previous state-of-the-art by a substantial margin of 2.1\% mIoU.

\end{abstract}

\vspace{-10pt}
\section{Introduction}
\vspace{-5pt}
\label{sec:intro}
\begin{figure}[!t]
    \centering
    \includegraphics[width=1.0\linewidth]{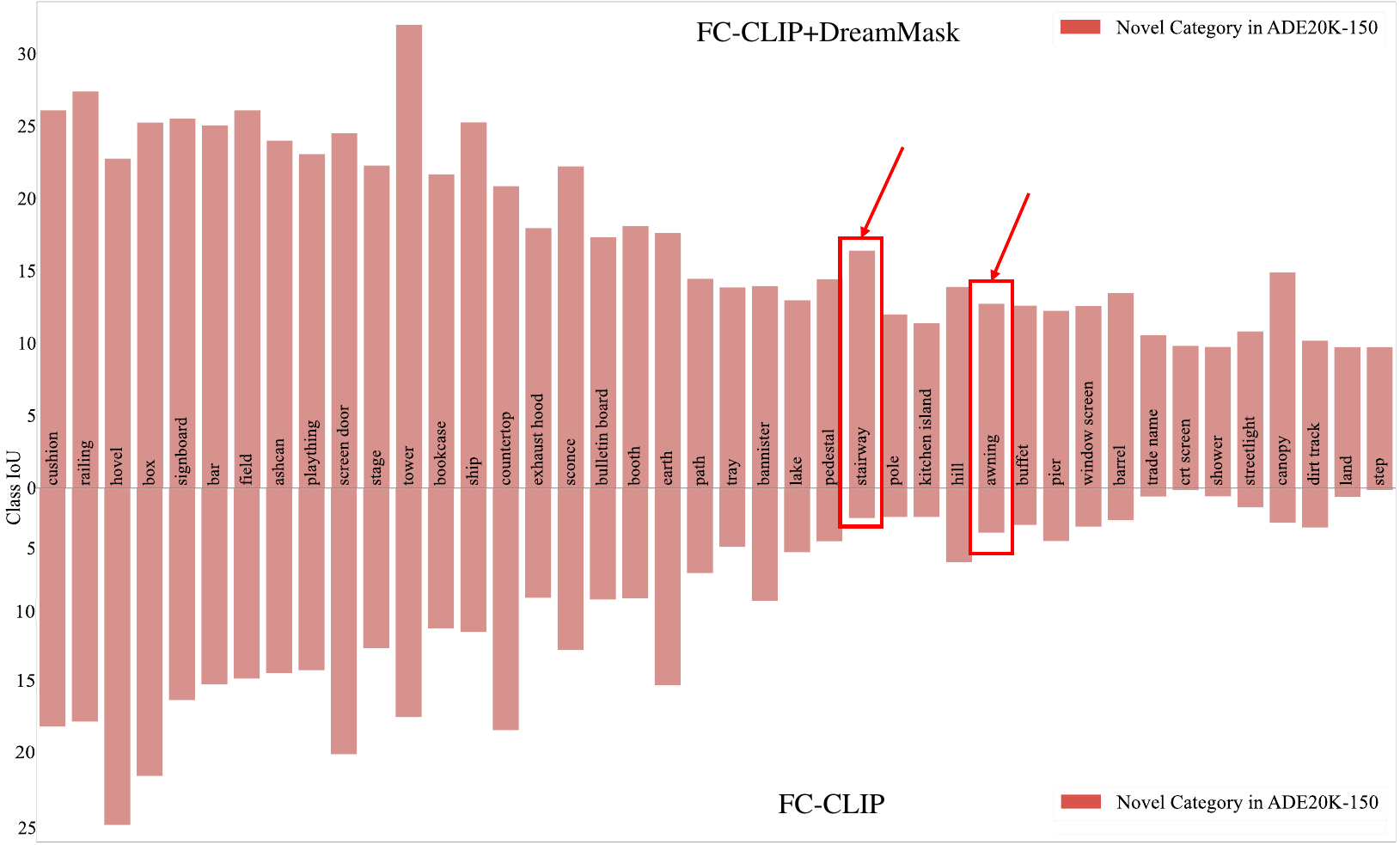}
    \caption{\textbf{Comparison of class IoU on novel classes in ADE20K~\cite{zhou2017scene}} (model trained on COCO~\cite{caesar2018coco}) between FC-CLIP~\cite{yu2023fcclip} and our FC-CLIP+DreamMask. Our method outperforms the baseline by a large margin on the accuracy of novel categories that have no similar semantics to the training categories of the COCO dataset. Detailed results are shown in the supplement.
    }
    \label{fig:motivation}
    \vspace{-4mm}
\end{figure}

Panoptic segmentation presents a challenging task in computer vision that classifies each pixel in an image into specific instances or stuff with category labels. Classical closed-vocabulary methods~\cite{kirillov2019panoptic, cheng2020panoptic,liu2019end} have achieved remarkable accomplishments by leveraging extensive, carefully annotated data. However, the acquisition of such large-scale data comes with high costs, limiting their practicality for novel categories. Moreover, these methods are restricted by fixed and predetermined label sets, making it difficult to associate arbitrary visual concepts and limiting their applicability in the real world.

Recently, open-vocabulary panoptic segmentation methods~\cite{chen2023open, xu2023open} address this issue by incorporating large-scale vision-language models, such as CLIP~\cite{radford2021learning}, which allows for almost any text query as the class definition.
However, since CLIP is trained on image-level contrastive objectives, it lacks the ability to conduct precise pixel-level recognition.
Therefore, directly combining it with existing fully-supervised segmentation models~\cite{kirillov2023segment} leads to inferior performance. 
To bridge this representation gap, previous studies typically fine-tune the model on the segmentation dataset, employing region-level or mask-level image-text representation alignment. 
Although these methods claim to improve the generalization ability, we find that
their advancements have primarily been concentrated on the classes overlapping with the training set, while only marginal progress has been made on novel categories. 
As shown in Fig.~\ref{fig:motivation}, the state-of-the-art method FC-CLIP~\cite{yu2023fcclip} shows notably inferior performance for novel categories, revealing limited generalization capability. 
This is a rather disappointing finding since recognizing the novel objects is the true purpose of open-vocabulary explorations.

This observation is however easily explained. The paradigm of modern large models tells us that the most effective way to improve generalization and achieve open-vocabulary segmentation is simply to expand the training set tremendously. This is easily confirmed by our experiments shown in Fig.~\ref{fig:motivation}. We empirically found that training with samples from larger-vocabulary datasets as a result of DreamMask contributes to better performance on novel classes (\textit{i.e.}, ``awning'': 1.47\% to 12.9\%, ``stairway'': 3.15\%
to 15.6\%).
Thus, we can conclude that enlarging the training vocabulary would be the most effective way to boost model performance, but segmentation data is highly costly to collect and annotate.  To this end, we can turn to using synthetic data to boost segmentation models that has been proven effective in previous works~\cite{sariyildiz2023fake,trabucco2023effective,azizi2023synthetic}.  However, previous explorations only tackle the close-vocabulary settings, where all categories are well-defined with provided masks.  
How to boost an open-vocabulary model with synthetic data is an unsolved question, where the difficulties lie in that the category name, the image, and the annotations are all unknown.

To tackle this challenge, we propose DreamMask, which synthesizes samples in novel categories to boost open-vocabulary segmentation. 
DreamMask consists of two core steps: novel sample synthesis, and imagination-aided training.   
In the first step, \emph{we empirically find that training with samples from randomly selected novel categories impairs the resulting performance due to the large domain gap with the initial training set (Tab.~\ref{tab:ablation})}. To address this issue, we leverage large language models (LLMs) to identify an extensive set of novel categories that have high correlations with the initial training categories. 
Subsequently, we randomly sample class names from the extended class set and leverage the LLMs again to generate reasonable layout descriptions. Then a well-trained layout-to-image diffusion model~\cite{feng2023layoutgpt} is used for context-aware synthesis. Following this, we utilize SAM~\cite{kirillov2023segment} and the generated layout to determine the masks of the generated samples, where we select the largest mask within each bounding box of the layout descriptions. 
We however observe that both unrealistic samples and samples with poor-quality annotations still remain at this point, for which we develop a multi-stage filtering mechanism to remove them.
In addition, previous studies~\cite{he2023synthetic,sariyildiz2023fake} have shown that training directly on synthetic images can lead to significant performance degradation due to the negative impact of inherent domain shift. To alleviate this problem during training in the second step, we propose a synthetic-real alignment strategy to minimize the feature discrepancy between realistic and synthetic samples. 

Our DreamMask method, as proposed, achieves exceptional performance for boosting open-vocabulary models. DreamMask is easily ``plug-and-play'', able to seamlessly integrate with existing methods, helping to improve their performance on novel categories (as shown in Fig.~\ref{fig:motivation}) without incurring any extra time cost during inference. To summarize, our main contributions  can be listed as follows:
\begin{itemize}

\item We present results from a thorough exploration to boost the open-vocabulary model with synthetic data, proposing DreamMask that expands the ``vocabulary" effectively via novel category imagination. An interesting insight here is that expanding to novel categories that are highly correlated to the training categories yields superior performance when compared to randomly selected categories.

\item We introduce a two-stage pipeline to make effective use of the synthetic samples. We first leverage LLMs for category name association and reasonable layout generation. Then the diffusion model are utilized for context-aware sample synthesis. For the latter, we design a synthetic-real alignment loss with class-wise prototypes, which mitigates the gap between synthetic and real samples.
\item We conduct experiments on various benchmarks and the results demonstrate that DreamMask achieves state-of-the-art performance, outperforming previous works by a substantial margin with a 2.1\% mIoU improvement. Interestingly, our experiments also show that synthetic data far outperforms data collected from the web, likely due to the much higher fidelity of the synthetic data. 




\end{itemize}

\section{Related Work}

\noindent\textbf{Open-vocabulary panoptic segmentation.} By unifying instance and semantic segmentation, open-vocabulary panoptic segmentation (OPS) targets at localizing objects from arbitrary categories during the inference stage, including categories that are included in the training data. Most prior works~\cite{wang2024open,jiao2024collaborative,qin2023freeseg,wu2023clipself,chen2023open,lan2024proxyclip,xu2023open, wang2023hierarchical,he2023primitive} either aim at generating class-agnostic mask or contribute to performing fine-grained image-text feature alignment. FreeSeg~\cite{qin2023freeseg} first proposes a unified framework by feeding the masked crops into the image encoder of CLIP. OPSNet~\cite{chen2023open} focuses on modulating mask/CLIP embedding. Inspired by ~\cite{rombach2022high}, ODISE~\cite{xu2023open} integrates text-to-image diffusion models to generate features with high correlation to semantic concepts. ProxyCLIP~\cite{lan2024proxyclip} leverages the spatial feature correspondence. However, though these methods have achieved substantial performance boost for the overall accuracy of segmentation, their improvement is generally observed only on the overlapping categories rather than novel ones. We posit that the training data owns a significant weight, and enlarging the training vocabulary would be the most effective way to boost model performance. Thus, inspired by recent advancements in image synthesis~\cite{dhariwal2021diffusion,dockhorn2022score,ho2020denoising} and LLMs~\cite{devlin2019bert}, we focus on enhancing the performance on novel categories by designing a novel sample imagination strategy, integrating LLMs and diffusion models.

\noindent\textbf{Vision-language pre-training.} Vision-language pre-training~\cite{chen2020uniter,lei2021less,li2020unicoder,radford2021learning} aims to encode images and languages jointly and perform multi-modal fusion. Early methods~\cite{chen2020uniter,lei2021less,li2020unicoder} involve fine-tuning features extracted from pre-trained detectors on downstream tasks using supervised signals derived from language. Recently, this field has witnessed the emergence of various methods~\cite{radford2021learning}, primarily driven by the rapid development of LLMs~\cite{devlin2018bert,brown2020language}. Among them, a milestone work CLIP~\cite{radford2021learning} stands out for its exceptional performance on zero-shot downstream tasks. Inspired by it, a myriad of studies has been proposed to leverage the knowledge embedded in the pre-trained CLIP model for downstream tasks, \ie, open-vocabulary segmentation.

\noindent\textbf{Synthetic images for training.} Multiple works have been proposed recently, which utilize either GANs~\cite{souly2017semi,antoniou2017data} or diffusion models~\cite{sariyildiz2023fake,trabucco2023effective,azizi2023synthetic, yang2023freemask} to generate samples for model training.~\cite{sariyildiz2023fake} attempts to replace the ImageNet dataset with pure synthetic images.~\cite{trabucco2023effective} focuses on editing available natural images with Stable Diffusion~\cite{dhariwal2021diffusion} to maintain image style. Similarly,~\cite{azizi2023synthetic} generates diverse photo-realistic samples given text prompts. To more effectively utilize synthetic data, FreeMask~\cite{yang2023freemask} designs a sample filtering criterion to suppress noisy synthetic images at both class and pixel levels. Even with all the advances made by these methods, there is still no method that can be used for open-vocabulary tasks since the novel categories that need to be synthesized are not known in advance. Moreover, they generally rely on masks available in the training set to generate masks for objects from novel categories, which will inevitably result in samples that are low in quality due to the domain shift. On the other hand, DreamMask leverages both LLMs and layout-to-diffusion models~\cite{qu2023layoutllm} to generate photo-realistic samples in a more context-aware manner for open-vocabulary panoptic segmentation methods.


\begin{figure*}[!t]
    \centering
    \includegraphics[width=1.0\linewidth]{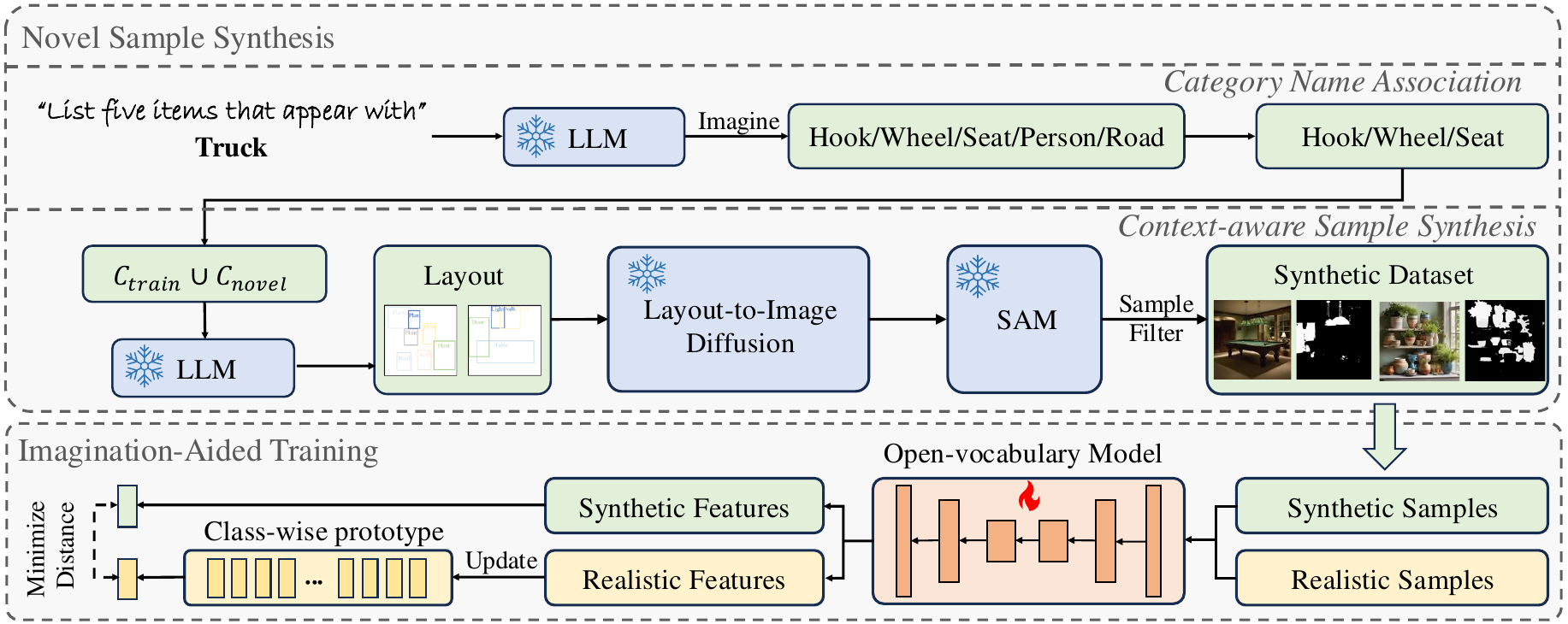}
    \vspace{-2mm}
    \caption{\textbf{The overall framework of DreamMask}, including \textbf{N}ovel \textbf{S}ample \textbf{S}ynthesis (NSS) and \textbf{I}magination \textbf{A}ided \textbf{T}raining (IAT). The former consists of \textbf{C}ategory \textbf{N}ame \textbf{A}ssociation and \textbf{C}ontext-aware Sample Synthesis (CSS). Specifically, CNA targets at extending novel class names with the powerful association abilities of LLMs, and CSS generates high-quality samples and corresponding pixel-level annotations with layout-to-image diffusion models and the SAM. Finally, in IAT, these samples are utilized to augment the training set and a synthetic-real alignment loss is introduced to alleviate the influence of domain shift by enclosing the representation between synthetic and realistic objects.
    }
    \vspace{-1mm}
    \label{fig:framework}
\end{figure*}


\section{Methodology}
\subsection{Overview}
\noindent\textbf{Problem definition.} Following ~\cite{xu2023open,chen2023open}, we train a model with a set of base training categories $\mathcal{C}_{train}$, which may be different from the categories in the test set $\mathcal{C}_{test}$, \ie $\mathcal{C}_{train}$$\neq$$\mathcal{C}_{test}$. $\mathcal{C}_{test}$ may contain samples from novel categories that are not in $\mathcal{C}_{train}$. We assume the binary panoptic segmentation mask together with its class label of each training sample are available. Note that only the category names are available during testing. The goal of panoptic segmentation is segmenting the image $\textbf{I}\in \mathbb{R}^{H \times W \times 3}$ into a set of masks with semantic and instance labels: $\left\{y^i\right\}_{i=1}^K=\left\{\left(m^i_r, c^i_r\right)\right\}_{i=1}^K$, where K masks can be denoted as ${m^i_r} \in \{0,1\}^{H \times W}$ and $c^i_r\in \mathcal{C}_{train}$\footnote{For all data-related symbols, we use the subscript of $r$/$s$ to denote the sample from realistic/synthetic views respectively.}. H/W are the height/width of the image respectively. The real and synthetic training sets are denoted as $\mathcal{D}_r$/$\mathcal{D}_s$ respectively. $f(\cdot)$ represents the open-vocabulary panoptic segmentation model, and the original segmentation loss of $f(\cdot)$ is denoted as $\mathcal{L}_{seg}$.

\noindent\textbf{Overall framework.} The overall framework of DreamMask is shown in Fig.~\ref{fig:framework}, which contains two core steps: \textbf{N}ovel \textbf{S}ample \textbf{S}ynthesis (NSS) and \textbf{I}magination \textbf{A}ided \textbf{T}raining (IAT). Specifically, in NSS, we first leverage the reasoning ability of LLMs to generate highly related novel categories $\mathcal{C}_{novel}$ to each class in $\mathcal{C}_{train}$. Then we randomly samples class names from $\mathcal{C}_{novel} \cup \mathcal{C}_{train}$ and leverages the LLMs again to generate reasonable layout descriptions. A pretrained layout-to-image diffusion model~\cite{feng2023layoutgpt} is used for context-aware sample synthesis. Afterward, the layout information and generated images are both fed into SAM to produce pixel-wise annotations for $\mathcal{C}_{train}$ and $\mathcal{C}_{novel}$, where a two-stage filtering mechanism is designed to retain high-quality samples. This is followed by IAT, where class-wise online-updating prototypes are constructed to alleviate the impact of domain shift by minimizing the feature discrepancy between real and synthetic objects.

\begin{figure}[!t]
    \centering
    \includegraphics[width=1.0\linewidth]{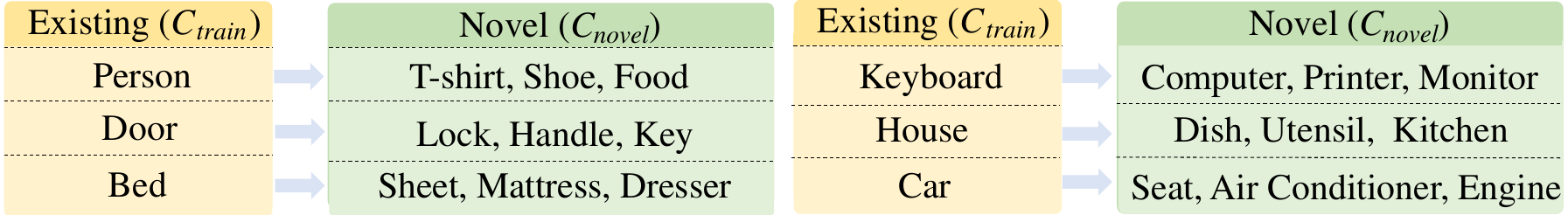}
    \vspace{-4mm}
    \caption{\textbf{Examples of extended novel categories.} Diverse class names can be generated in category name association with the powerful reasoning abilities of LLMs.   
    }
    \vspace{-1mm}
    \label{fig:extend}
\end{figure}

\begin{figure}[!t]
    \centering
    \includegraphics[width=1.0\linewidth]{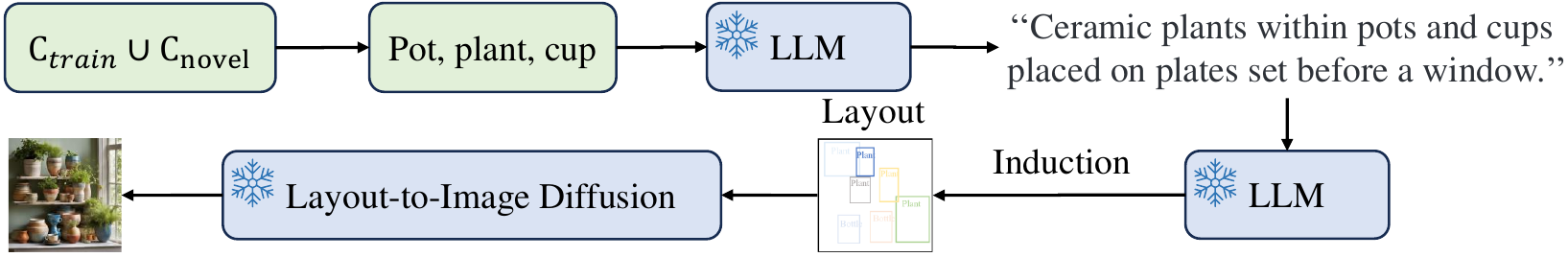}
    \caption{\textbf{Context-aware generation.} The class names are sampled from $\mathcal{C}_{novel}$ $\cup$ $\mathcal{C}_{train}$ and then fed into the LLMs to produce coarse-grained layout descriptions. Then such results are fed into the LLMs again for fine-grained visual planning. A layout-to-image diffusion model~\cite{feng2023layoutgpt} is followed for high-quality synthesis.
    }
\vspace{-1mm}
\label{fig:context}
\end{figure}

\subsection{Novel Sample Synthesis}
\noindent\textbf{Category name association (CNA).} In the CNA step, in light of advances in LLMs~\cite{devlin2018bert,brown2020language}, we aim to leverage their reasoning capabilities to extend the scope of novel categories for training. Specifically, the initial training categories $\mathcal{C}_{train}$ are first utilized as input to the LLM along with text prompts to generate an extensive set of highly related novel class names. Subsequently, we remove wrong items (\ie non-noun names) as well as filter out any generated category names that already exist in the training set or are semantically similar to categories in $\mathcal{C}_{train}$. Examples of the generated novel categories are shown in Fig.~\ref{fig:extend}. Due to the unstable outputs of LLMs, we perform the operation five times and select repeating class names as the final reliable categories, denoted as $\mathcal{C}_{novel}$ for convenience.

\noindent\textbf{Context-aware sample synthesis (CSS).} 
Given the novel class names generated by the CNA step, we leverage the powerful scene planning ability of LLMs. Specifically, as shown in Fig.~\ref{fig:context}, we randomly select classes from $\mathcal{C}_{novel}$ $\cup$ $\mathcal{C}_{train}$ and leverage the LLMs to perform coarse-grained layout planning for the selected classes. Then the generated descriptions are transformed into fine-grained layout information through text-to-layout induction~\cite{qu2023layoutllm,feng2023layoutgpt} and a pre-trained layout-to-image diffusion model~\cite{feng2023layoutgpt} is utilized to synthesize samples that are faithful to the generated layout and scene descriptions. Subsequently, these samples are provided as inputs to SAM~\cite{kirillov2023segment} with their layout information (\ie, bounding boxes) as prompts to generate corresponding class-agnostic masks. Given that the synthetic layouts guarantee the presence of objects from a single category in the area within each bounding box of each synthetic image, obtaining pixel-wise annotations for each object becomes straightforward with the generated masks.

\noindent\textbf{Score-based selection.} Although visually appealing, using all synthetic objects for training without discrimination may lead to a performance decline due to the variability in generation and annotation quality. 
To ensure we select only the highest quality synthetic samples, we attempt to perform filtering based on both generation and annotation quality. 

First, 
tapping into the ability of the CLIP model~\cite{radford2021learning} in distinguishing between real and synthetic images, we adopt a CLIP score guided filtering approach to exclude samples with artifacts from training. Specifically, we feed objects cropped with each bounding box and its class name into the image and text encoder of the CLIP model respectively, and calculate the similarity score between the extracted features. Then the mean similarity score within each image is adopted as the final CLIP score, with a larger score indicating higher quality of the synthetic object. Using the score as the selection criterion, we retain samples with scores that are larger than the average of the whole synthetic dataset for subsequent training. 

Afterwards, to address the sometimes hit-and-miss quality of the annotations generated by the SAM, we utilize the uncertainties of the pixel-wise predictions for each object as guidance to filter out objects with low-quality annotations. However, we empirically find that it will induce severely imbalanced distributions in the samples that have been retained when directly filtering all the samples regardless of the class-wise discrepancy. Thus we opt for a more class-sensitive filtering approach. Specifically, we first calculate the mean pixel-wise uncertainty for the predictions of each retained object, where the process can be formulated as:
\vspace{-2mm}
\begin{equation}
S_i=\sum_{h w}^{H W}\left[\mathbbm{1}\left(\mathbf{M}_{h w}^i=j\right) \times (\mathbbm{1}-\mathbf{P}_{h w}^i)\right] / \sum_{h w}^{H W} \mathbbm{1}\left(\mathbf{M}_{h w}^i=j\right),
\end{equation}

\noindent where $\mathbf{P}_i$ and $\mathbf{S}_i$ denote the pixel-wise predictions and corresponding mean uncertainty of \textit{i}-th objects respectively. Then we select top-\textit{n} objects with the lowest prediction uncertainty to construct the final synthetic set. Despite the simplicity of such a two-stage filtering criterion, it significantly boosts the generalization performance of the target model. For convenience, the filtered synthetic set is denoted as $D_s = \{(x^i_s, m^i_s)\}_{i=1}^{n_s}$, where $n_s$ is the number of synthetic images.

\subsection{Training with Synthetic Data}
\noindent\textbf{Synthetic-real alignment.} With the high-quality synthetic sample set from NSS that makes up $\mathcal{C}_{novel}$, we can directly utilize them and $\mathcal{C}_{train}$ for model training. However, due to the domain shift between $\mathcal{C}_{train}$ and $\mathcal{C}_{novel}$, we observe that the resulting models are biased towards synthetic images, failing to generalize well on real images in the inference stage. To address this issue, we design a loss term that serves to align the representations of real and synthetic objects by constructing online-updating class-wise memory banks for each category in $\mathcal{C}_{train}$. Specifically, the memory bank of the features of real objects from \textit{p}-th category in $\mathcal{C}_{train}$ is denoted as:
\begin{equation}
\begin{aligned}
\mathcal{M}^p_r=\left[\begin{array}{llll}
\mathcal{F}^{0}_r, & \mathcal{F}^{1}_r, & \ldots, & \mathcal{F}^{\beta}_r
\end{array}\right]^p,
\end{aligned}
\end{equation}
\noindent where $\beta$ is the length of the memory bank. The most/least recently used feature will be enqueued/dequeued in $\mathcal{M}^p_r$ during the training process. $\mathcal{F}_r\in \mathbb{R}^{L}$ is the feature generated by $f(\cdot)$, where $L$ is the feature length. Thus, the feature prototype of real objects from the \textit{p}-th category in $\mathcal{C}_{train}$ can be formulated as $\overline{\mathcal{M}^p_r}$, while we denote the feature of synthetic objects from the same category as $\mathcal{F}_s^p$. The synthetic-real alignment loss is then calculated based on the cosine similarity between $\mathcal{F}_s^p$ and $\overline{\mathcal{M}^p_r}$:
\begin{equation}
    \mathcal{L}_{sra} = 1-\frac{\mathcal{F}_s^p \cdot \overline{\mathcal{M}^p_r}}{||\mathcal{F}_s^p||_2 \cdot ||\overline{\mathcal{M}^p_r}||_2}.
    \label{eq:sra}
\end{equation}
 The overall loss function for DreamMask 
 is denoted as:
\begin{equation}
    \mathcal{L} = \mathcal{L}_{seg} + \lambda\mathcal{L}_{sra},
    \label{eq:overall}
\end{equation}
\noindent where $\lambda$ is a balance parameter between $\mathcal{L}_{seg}$ and $\mathcal{L}_{sra}$. $\mathcal{L}_{seg}$ denotes the original loss of the combined method and DreamMask only introduces $\mathcal{L}_{sra}$ for training. 

\noindent\textbf{Overall algorithm.} Putting this all together, Algorithm~\ref{pseudoalgorithm} outlines the proposed DreamMask for open-vocabulary panoptic segmentation. Firstly, in NSS, $\mathcal{C}_{train}$ is fed into the LLM to leverage its powerful association ability to produce extensive novel categories $\mathcal{C}_{novel}$. Then class names are randomly sampled from $\mathcal{C}_{train}$ $\cup$ $\mathcal{C}_{novel}$ as inputs to the LLMs again to produce detailed layout descriptions with text-to-layout induction. Next, a pre-trained layout-to-image diffusion model is utilized in NSS to synthesize highly faithful samples. A two-stage filtering strategy follows to retain high-quality samples in terms of generation and annotation quality based on both CLIP scores and prediction uncertainty. Afterward, in IAT, the retained synthetic samples are employed to augment the training set. A synthetic-real alignment loss is designed to further mitigate the influence of domain shift by enclosing features between synthetic and real images with class-wise online-updating prototypes. Finally, DreamMask can be easily plugged into existing works to boost their performance on novel categories without incurring huge labeling cost.   

\begin{algorithm}[!t] 
    \footnotesize
    \caption{The proposed DreamMask strategy}
    \begin{algorithmic}[1]
    \renewcommand{\algorithmicrequire}{\textbf{Input:}}
    \REQUIRE Training categories $\mathcal{C}_{train}$, open-vocabulary panoptic segmentation model $f(\cdot)$, text-to-image diffusion model $F(\cdot)$, max iteration number $T$, batch size \textit{b}.  
    \renewcommand{\algorithmicrequire}{ \textbf{Procedure:}}
    \REQUIRE
    \STATE Feed $\mathcal{C}_{train}$ into LLMs to generate $\mathcal{C}_{novel}$.
    \STATE Feed classes sampled from $\mathcal{C}_{train}$$\cup$ $\mathcal{C}_{novel}$ into LLMs to obtain layout descriptions.
    \STATE Feed layout descriptions to $F(\cdot)$ to generate samples.
    \STATE Filter high-quality samples based on CLIP scores.
    \STATE Generate masks with retained samples and layout descriptions.
    \STATE Filter high-quality annotations based on prediction uncertainty to form $\mathcal{D}_s$.
    \FOR{$i = 1$ to $T$}
    \STATE Sample $n \times b$ objects from $\mathcal{D}_s$.
    \STATE Sample \textit{b} images from $\mathcal{D}_r$ $\cup$ $\mathcal{D}_s$.
    \STATE Calculate $\mathcal{L}$ based on Eq.~\ref{eq:overall}.
    \STATE Update $\mathcal{M}_r$ with features of realistic objects.
    \STATE Update parameters of $f(\cdot)$ based on $\mathcal{L}$ in backward process.
    \ENDFOR
    \renewcommand{\algorithmicensure}{\textbf{Output:}}
    \ENSURE The final model $f(\cdot)$.
    \end{algorithmic}  
    \label{pseudoalgorithm}
     
\end{algorithm}

\section{Experiments}

\subsection{Experimental Setup}




\noindent\textbf{Implementation details.} The LayoutGPT~\cite{feng2023layoutgpt} is used as the layout-to-image generation model and the SAM~\cite{kirillov2023segment} with a ViT-H backbone is utilized for mask generation. For the model training, we follow the same setting with ~\cite{yu2023fcclip} and include no special modification, where the model is trained for 50 epochs on the COCO~\cite{lin2014microsoft} panoptic segmentation training set. A learning rate of $1\times10^{-4}$ is employed and decayed in a multi-step schedule. The batch size is set as 16 for training and the image in $\mathcal{D}_r$ and $\mathcal{D}_s$ is resized to $1024 \times 1024$. We adopt GPT-3.5 as LLMs and feed the prompt in Fig.~\ref{fig:framework} to it five times, selecting overlapped class names as the final reliable classes. Details of the inference and evaluation protocols are provided in the Appendix.

\begin{table*}[!t]
\caption{
    \label{tab:semantic}
    \textbf{Open-vocabulary semantic segmentation performance.}
    When combing with FC-CLIP, our DreamMask surpasses FC-CLIP and previous methods across all the benchmarks. * denotes utilizing Swin-B~\cite{liu2021swin} as the backbone network.
}
\resizebox{\linewidth}{!}{

\begin{tabular}{l|c|cccccc}

\toprule
\multirow{2}*{Method}  &  \multirow{2}*{Training dataset}   & \multicolumn{6}{c}{mIoU}\\

                           &     & A-847         & PC-459     & A-150         & PC-59         & PAS-21  & PAS-20        \\
\hline
ZS3Net~\cite{bucher2019zero}     & Pascal VOC~\cite{everingham2010pascal}           & -             & -             & -             & 19.4          & 38.3       & -       \\
CAT-Seg~\cite{cho2023cat}     & COCO Stuff~\cite{caesar2018coco}   &8.4        & 16.6     &27.2        &57.5         &-       &93.7      \\ 
DeOP~\cite{Han_2023_ICCV}     & COCO Stuff~\cite{caesar2018coco}     & 7.1       &9.4      &22.9        &       48.8  &    -   & 91.7      \\
\hline
GroupViT~\cite{xu2022groupvit}     & GCC~\cite{sharma2018conceptual}+YFCC~\cite{thomee2016yfcc100m}     & 4.3           & 4.9           & 10.6          & 25.9          & 50.7        & 52.3     \\
\hline
LSeg+~\cite{li2022language,ghiasi2022scaling}        & COCO Stuff~\cite{caesar2018coco}              & 3.8           & 7.8           & 18.0          & 46.5          & -         & -  
\\
OVSeg~\cite{liang2022open}      & COCO Stuff~\cite{caesar2018coco}           & 9.0           & 12.4           & 29.6          & 55.7          & -         & 94.5      \\
SAN~\cite{xu2023side}      & COCO Stuff~\cite{caesar2018coco}           & 13.7           & 17.1           & 33.3          & 60.2          & -        & 95.5        \\

\hline
OpenSeg~\cite{ghiasi2022scaling}      & COCO Panoptic + COCO Caption          & 6.3           & 9.0           & 21.1          & 42.1          & -         & -       \\
OVSeg*~\cite{liang2022open}     & COCO Stuff~\cite{caesar2018coco} + COCO Caption     & 9.0       &12.4      &  29.6      &  55.7       &    -   &94.5      \\
CELoss+OVSeg~\cite{dao2023class}     & COCO Stuff~\cite{caesar2018coco} + COCO Caption    &  9.7      &12.6      &   29.9     &55.6         & -      &91.8      \\

ODISE~\cite{xu2023open} (caption)        & COCO Panoptic + COCO Caption      & 11.0 & 13.8 & 28.7 & 55.3 & 82.7   & -  \\
\hline
MAFT~\cite{jiao2023learning}       & COCO Panoptic              & 13.1           & 17.0          & 34.4         & 57.5          & -         & 93.0    \\

\hline
ODISE~\cite{xu2023open}        & COCO Panoptic            & 11.1 & 14.5 & 29.9 & 57.3 & 84.6   & -  \\
\rowcolor{mygray}
ODISE+DreamMask (Ours)        & COCO Panoptic  & 12.7  &  16.0 & 31.4 & 58.1  &\textbf{85.0}    & -   \\ \hline

FC-CLIP~\cite{yu2023fcclip} (Baseline)          & COCO Panoptic      & 14.8 & 18.2 & 34.1 & 58.4 & 81.8   & 95.4  \\ 
\rowcolor{mygray}
FC-CLIP+DreamMask (Ours)        & COCO Panoptic  & \textbf{16.8}  & {20.5}  & {37.4} & {59.2}  &{82.2}     & 95.7   \\ \hline

MAFT+~\cite{jiao2024collaborative}        & COCO Panoptic            & 15.1 & 21.6 & 36.1 & 59.4 & -   & 96.5  \\
\rowcolor{mygray}
MAFT++DreamMask (Ours)        & COCO Panoptic  &  \textbf{16.8} & \textbf{22.8}  & \textbf{38.2} & \textbf{60.6}  &-    &  \textbf{96.8}  \\ 

\bottomrule

\end{tabular}}

\end{table*}
\begin{table}[!t]
\tablestyle{5.5pt}{1}
\centering
\caption{
    \label{tab:panoptic_similar}
    \textbf{Panoptic segmentation performance} for open- and close-vocabulary settings on ADE20K and COCO.
    DreamMask can boost the performance of open-vocabulary panoptic segmentation for all the methods. And comparable accuracy can be achieved on close-vocabulary tasks by it as well.
}
\resizebox{1.0\linewidth}{!}{
\begin{tabular}{l|ccc|ccc}
\toprule

{\multirow{2}{*}{Methods}} & \multicolumn{3}{c|}{ADE20K}                     & \multicolumn{3}{c}{COCO}                      \\

                      & PQ            & AP   & mIoU     & PQ     & AP     & mIoU          \\
\hline
MaskCLIP~\cite{ding2022open}   & 15.1  & 6.0             & 23.7      & -     & -  & -           \\
FreeSeg~\cite{qin2023freeseg}   & 16.3   & 6.5             & 24.6         & -  & -  & -           \\
ODISE~\cite{xu2023open} (caption)  & 23.4 & 13.9  & 28.7 & 45.6 & 38.4  & 52.4  \\ \hline 
ODISE~\cite{xu2023open}   & 22.6    & 14.4      & 29.9  & 55.4 & 46.0   & 65.2 \\
\rowcolor{mygray}
ODISE+DreamMask (Ours)   & 24.2    & 15.3      & 31.4  & {57.2} & {46.4}   & \textbf{65.4} \\ \hline
FC-CLIP~\cite{yu2023fcclip}    & 26.8 & 16.8 & 34.1  & 54.4 & 44.6  & 63.7 \\
\rowcolor{mygray}
FC-CLIP+DreamMask (Ours)& {28.1} & {18.7} & {37.4}  & 55.0  & 44.9 & 63.9 \\ \hline

MAFT+~\cite{jiao2024collaborative}    &28.7  & 15.1 & 36.1   &57.3  & 45.8  & 64.5 \\
\rowcolor{mygray}
MAFT++DreamMask (Ours)& \textbf{30.1} & \textbf{16.8} & \textbf{38.2}  & \textbf{57.8}  & \textbf{46.5} & 65.3 \\
\bottomrule
\end{tabular}}

\end{table}

%


\subsection{Experimental Results}

\begin{table}[!t]
    \centering
    \caption{
    \label{tab:semanticnovel}
    \textbf{Performance on novel categories} that have no similar semantics to the training categories in COCO for open-vocabulary semantic segmentation. DreamMask significantly boosts the performance of all the methods across three benchmarks. 
    }
\tablestyle{5.5pt}{1}
\resizebox{0.9\linewidth}{!}{
\begin{tabular}{l|ccc}
\toprule
\multirow{2}*{Method}  & \multicolumn{3}{c}{mIoU on Novel Classes}                                   \\

 & A-847         & PC-459     & A-150     \\
\hline
ODISE~\cite{xu2023open}                & 2.0  &1.5  &9.0   \\
\rowcolor{mygray}
ODISE+DreamMask (Ours)        & 5.4  & 6.9  & 16.7  \\ \hline
FC-CLIP~\cite{yu2023fcclip}     &  2.5 & 1.7  & 8.5  \\
\rowcolor{mygray}
FC-CLIP+DreamMask (Ours)      &  6.8 &  8.6 & 21.8   \\ \hline

MAFT+~\cite{jiao2024collaborative}     & 4.6  & 3.8  &  10.1 \\
\rowcolor{mygray}
MAFT++DreamMask (Ours)      & 7.5   & 9.3  & 22.5   \\ 
\bottomrule
\end{tabular}}
\end{table}

\begin{table}
\centering
\caption{
    \label{tab:component}
    \textbf{Ablation study on NSS and IAT.} NSS/IAT denote novel sample synthesis/imagination-aided training. FC-CLIP is the baseline. Both NSS/IAT contribute to the best accuracy.
}
\resizebox{\linewidth}{!}{
\tablestyle{16pt}{1}
\begin{tabular}{l|ccc}
\toprule
{\multirow{2}{*}{Methods}} & \multicolumn{3}{c}{mIoU}  \\

& A-150  & A-847  & PC-459 \\
\hline
Baseline   & {34.1} & {14.8} & {18.2}  \\
\hline
+ NSS & 35.7  & 15.6  & 19.0    \\ 
\rowcolor{mygray}

+ NSS \& IAT & \textbf{37.4} & \textbf{16.8}  & \textbf{20.5}    \\  
\bottomrule
\end{tabular}}
\end{table}

\begin{table}
\centering
\caption{
    \label{tab:ablation}
    \textbf{Investigation on the augmentation manners}, including \textbf{c}opy-\textbf{p}asting, MosaicFusion~\cite{xie2023mosaicfusion} (MF), and our context-aware generation strategy. ‘Random’ denotes randomly selecting the same number of new classes as our extended set based on the LLM. ‘Web-crawl’ is training with the same number of web-crawled data as our NSS. FC-CLIP is adopted as the baseline. 
}
\resizebox{\linewidth}{!}{
\tablestyle{16pt}{1}
\begin{tabular}{l|ccc}
\toprule
{\multirow{2}{*}{Methods}} & \multicolumn{3}{c}{mIoU}  \\

& A-150  & A-847  & PC-459  \\
\hline
Baseline   & {34.1} & {14.8} & {18.2}  \\
\hline
+CP  & 32.4  & 12.1  & 16.3   \\ 
+MF  &34.6 & {15.3} & {18.4} \\ 
\hline
+ Web-crawl    &34.6  & 14.9 & 19.0   \\
+ Random    &32.3  & 13.7 & 17.6   \\
\hline
\rowcolor{mygray}
+NSS (Ours)  & \textbf{37.4} & \textbf{16.8}  & \textbf{20.5}    \\  

\bottomrule
\end{tabular}}
\end{table}

\noindent\textbf{Open-vocabulary semantic segmentation.} The comparison between our DreamMask and previous works on open-vocabulary semantic segmentation is shown in Tab.~\ref{tab:semantic}. Following ~\cite{yu2023fcclip}, we present the mIoU results for five settings, including: (a) A-150: consists of 150 common classes in ADE20K~\cite{zhou2017scene}; (b) A-847: includes all 847 classes from the ADE20K dataset~\cite{zhou2017scene}; (c) PC-59: contains 59 common classes from the Pascal Context dataset~\cite{everingham2010pascal}; (d) PC-459: encompasses the full 459 classes from the Pascal Context dataset~\cite{everingham2010pascal};  (e) The classic Pascal VOC dataset~\cite{everingham2010pascal}, with 20 foreground classes and 1 background class (PAS-21) and (f)PAS-20: contains only 20 foreground classes in PAS-21. The results show that DreamMask can significantly boost their performance across all the benchmarks. Specifically, DreamMask boosts MAFT+~\cite{jiao2024collaborative} by 1.7\% and 2.1\% on A-847 and A-150 respectively, setting a new state-of-the-art record. FC-CLIP+DreamMask outperforms FC-CLIP by a large margin: 3.3\% mIoU on A-150, 2.0\% mIoU on A-847, 2.0\% mIoU on PC-459. And it shows superior performance over methods that utilize image caption for supervisory signals or utilize the COCO-Stuff dataset for training. Though a smaller relative improvement is achieved by DreamMask on PAS-21 and PAS-20, it mainly stems from the fact that the categories in these two datasets are all included in $C_{train}$, limiting DreamMask from fully demonstrating its power. Additionally, similar performance boost can be observed on ODISE as well when it combines with DreamMask. All these results demonstrate the great generality of DreamMask, and that it can be seamlessly integrated with existing works as a plug-in-play module to boost their performance on open-vocabulary segmentation tasks.

\noindent\textbf{Open-vocabulary panoptic segmentation.} We evaluate DreamMask by combining it with three recent models, MAFT+~\cite{jiao2024collaborative}, FC-CLIP~\cite{yu2023fcclip} and ODISE~\cite{karazija2023diffusion}. Tab.~\ref{tab:panoptic_similar} shows that MAFT++DreamMask achieves state-of-the-art performance. FC-CLIP+DreamMask surpasses FC-CLIP by a margin of 1.3\% in PQ, 1.9\% in AP, and 3.3\% in mIoU on ADE20K-150. Furthermore, to showcase the superiority of DreamMask in a more intuitive way, we also report the performance of three baselines on novel categories in the test set in Tab.~\ref{tab:semanticnovel}. The results demonstrate that DreamMask significantly improves accuracy on novel categories of existing methods.

\noindent\textbf{{Close-vocabulary panoptic segmentation.}} We also report the performance under the close-vocabulary panoptic segmentation settings. We showcase the outcome on the COCO training dataset in Tab.~\ref{tab:panoptic_similar}. The results demonstrate that when combined with our DreamMask, existing methods achieve comparable performance to the baseline. This signifies the effectiveness of our approach in bridging the gap between real and synthetic objects representations.

\begin{figure}[!t]
    \centering
    \includegraphics[width=1.0\linewidth]{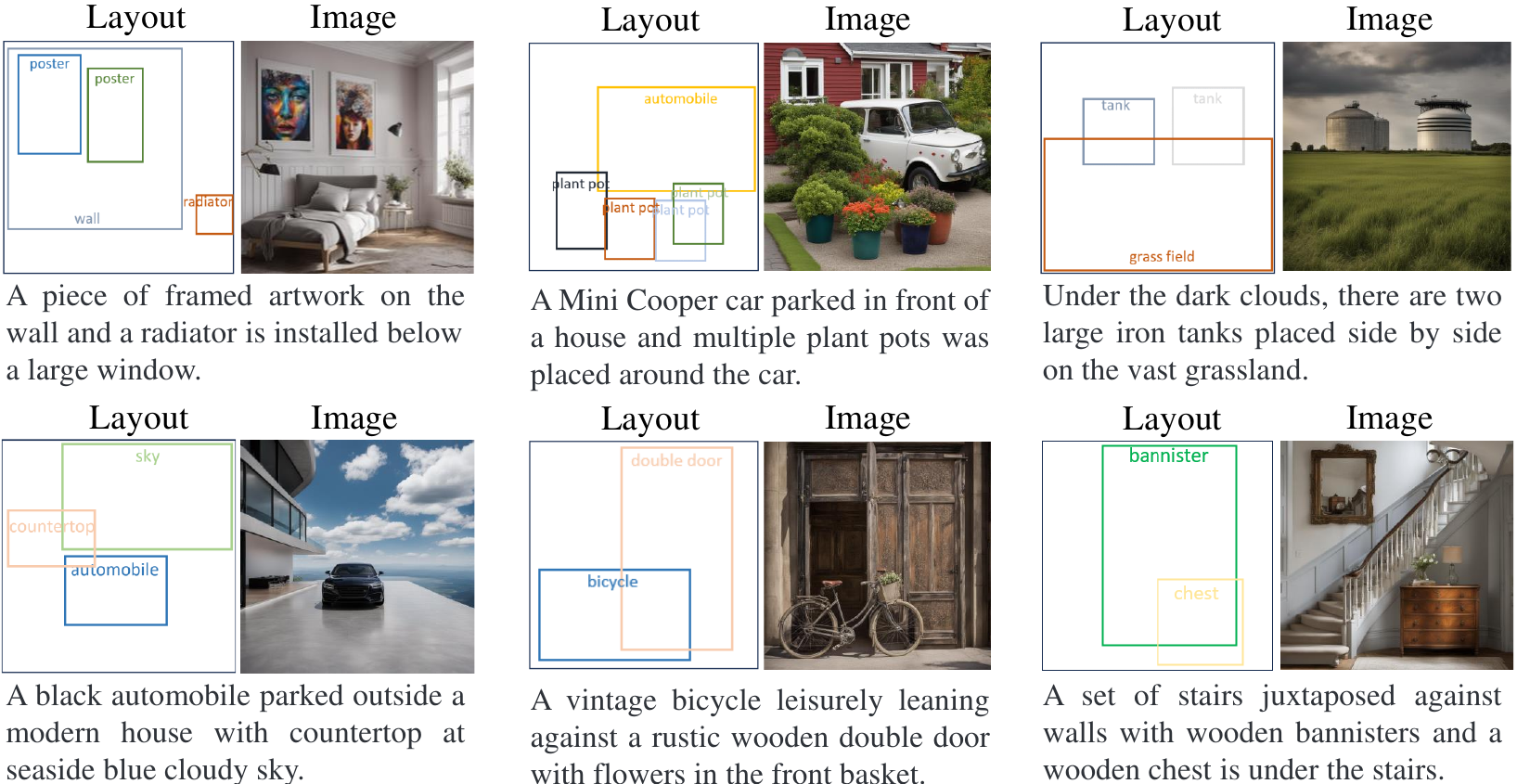}
        \vspace{-2mm}
    \caption{Examples of the text descriptions, layouts and corresponding synthetic samples. The LLMs can effectively help generate samples with realistic layouts.
    }
\vspace{-2mm}
    
    \label{fig:layout}
\end{figure}

\begin{figure}[!t]
    \centering
    \includegraphics[width=1.0\linewidth]{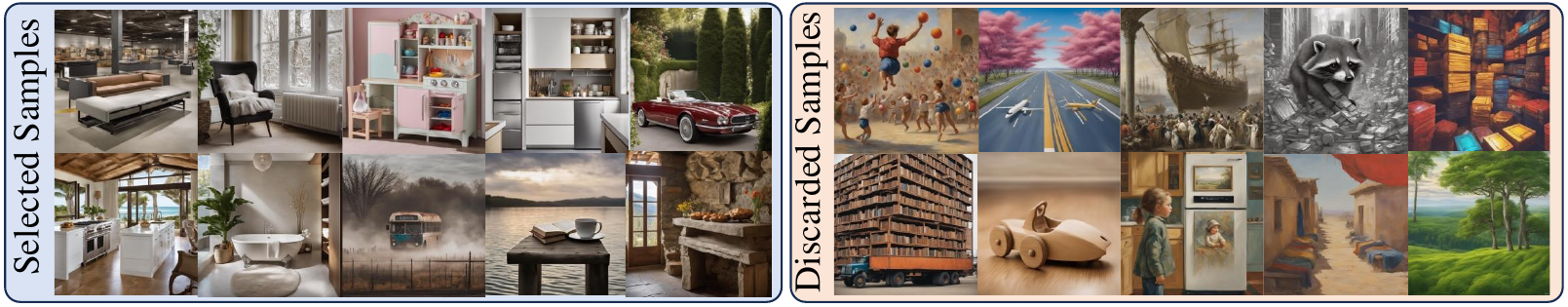}
        \vspace{-2mm}
    \caption{Examples of the selected/discarded samples. The proposed CLIP-score based
filter strategy can effectively retain high-quality samples with more realistic styles.
    }
\vspace{-2mm}
    
    \label{fig:discard}
\end{figure}

\vspace{-1mm}
\subsection{Ablation Studies}
\vspace{-1mm}

\noindent\textbf{Component analysis.} We conduct experiments under three settings to provide a detailed analysis on the effect of two main components: Novel Sample Synthesis (NSS) and Imagination-aided Training (IAT). As shown in Tab.~\ref{tab:component}, ‘+NSS’ denotes training without synthetic-real alignment, while ‘+IAT’ denotes training with all the generated samples without selection. Clearly, NSS can bring a significant performance boost to the baseline, and it induces the best performance when training with both modules. Moreover, to further verify the effectiveness of our filter strategy, we provide the visualization of selected and discarded synthetic samples in the Appendix as well, which shows that our method can effectively retain more realistic samples with high-quality pixel-wise annotations.

\noindent\textbf{Filtering strategy.} To verify the effectiveness of our filter strategy, we visualize the selected and discarded samples in Fig.~\ref{fig:discard}. The results demonstrate that it can effectively retain high-quality samples from extensive synthetic data. 

\noindent\textbf{Context-aware \textit{vs.} context-unaware synthesis.} Here we conduct a detailed investigation to verify the superiority of our context-aware synthesis scheme over context-unaware ones. Specifically, we augment the training set in two context-unaware manners respectively, including copy-pasting and MosaicFusion~\cite{xie2023mosaicfusion}. For copy-pasting, we directly synthesize samples with single object from $\mathcal{C}_{train}$ and $\mathcal{C}_{novel}$ with pre-trained text-to-image diffusion models. Then we generate the corresponding masks with the SAM~\cite{kirillov2023segment} by selecting the mask with the largest area. Afterward, we copy-paste the objects cropped with masks on the realistic images of the training set and train the model with the augmented samples. A similar synthetic-to-real alignment strategy is adopted for both copy-pasting and MosaicFusion~\cite{xie2023mosaicfusion} for fair comparison. And as shown in Tab.~\ref{tab:ablation}, our NSS outperforms the context-unaware ones by a large margin by synthesizing samples with realistic contexts.

\noindent\textbf{Analysis on CNA.} To further verify the necessity of the CNA, we randomly generate the same number of new categories based on the LLM and conduct sample synthesis with these classes. Specifically, as shown in Tab.~\ref{tab:ablation}, training with samples from random classes slightly degrades the performance due to the large domain gap between the initial class set and the randomly generated one. By contrast, our CNA significantly boosts the accuracy and outperforms ‘Random’ by a large margin of 5.1\% on A-150.

\noindent\textbf{Comparison with web-crawled data.} Here we provide comparisons with results of training with web-crawled data to further demonstrate the advantage of synthetic data. Specifically, 
we crawl an equal number of images as those generated by NSS from Google based on the categories selected by CNA, and then use SAM~\cite{kirillov2023segment} to generate the corresponding masks. Tab.~\ref{tab:ablation} shows that such a scheme only achieves nearly the same performance as the baseline due to the negative impact of noisy labels produced by SAM~\cite{kirillov2023segment}.

    

\begin{figure}[!t]
    \centering
    \includegraphics[width=1.0\linewidth]{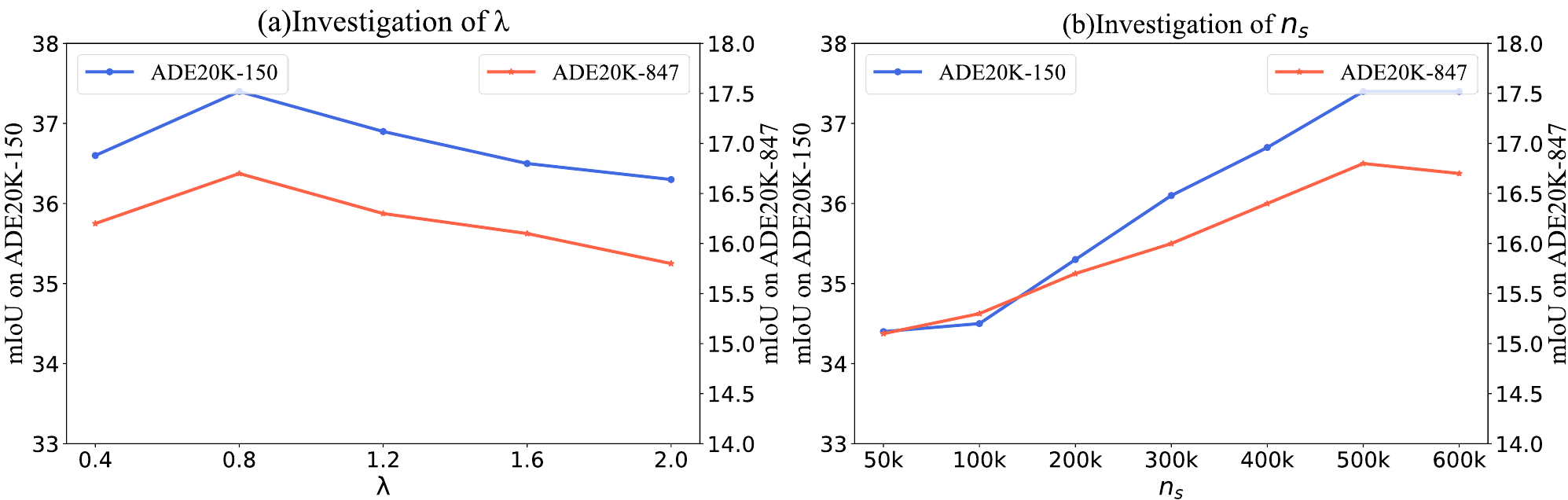}
    \vspace{-3mm}
    \caption{Investigation of (a) the balance parameter $\lambda$, (b) the sample number $\textit{n}_s$ in each novel category on the performance of FC-CLIP+DreamMask. Experiments are conducted on ADE20k-150 and ADE-847. 
    }
\vspace{-3mm}
    
    \label{fig:sensitivity}
\end{figure}

\begin{figure*}[!t]
    \centering
    
    \includegraphics[width=1.0\linewidth]{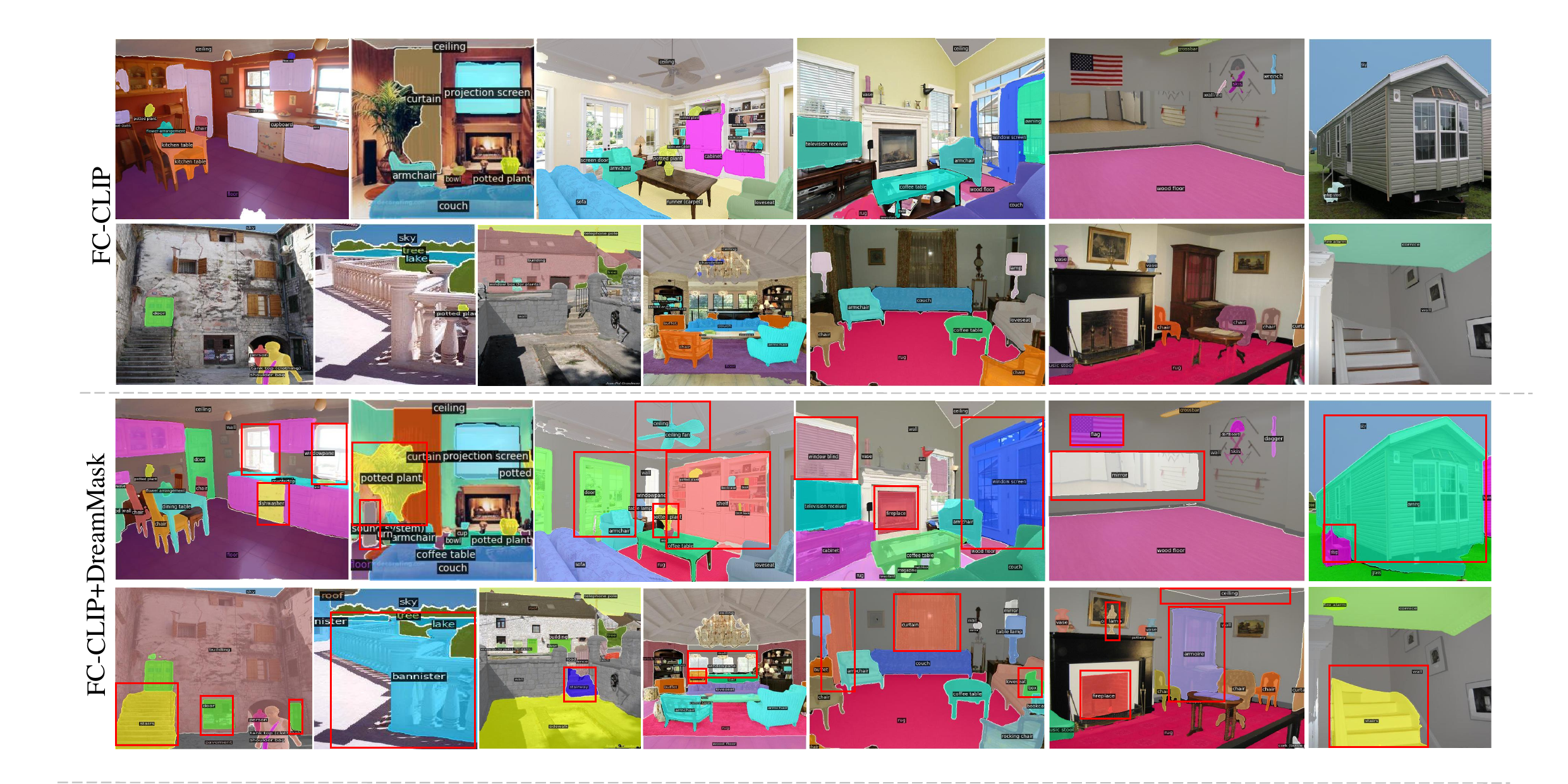}
    \vspace{-4mm}
    \caption{\textbf{Qualitative visualization of FC-CLIP and FC-CLIP+DreamMask} on ADE20K validation set. The red boxes point out the objects that are successfully segmented in FC-CLIP+DreamMask while fail to be recognized by FC-CLIP. The results demonstrate that DreamMask can effectively help FC-CLIP localize objects from various categories, especially the novel ones. “Windowpane”, “stairway”, “ceiling fan”, “bannister”, “stairs”, etc., are novel categories that are not annotated in COCO. 
    }
    \label{fig:qualitative}
\end{figure*}
\noindent\textbf{Parameter sensitivity.} We conduct investigation on two parameters in DreamMask: (a) the balance parameter $\lambda$, (b) the sample number $\textit{n}_s$ in each novel category on the performance of FC-CLIP+DreamMask. Specifically, as depicted in Fig.~\ref{fig:sensitivity} (a), DreamMask achieves the best performance when $\lambda$=0.8 and the accuracy gradually decreases when $\lambda \textgreater$0.8. Such a phenomenon may be accounted for by the fact that larger $\mathcal{L}_{sra}$ will induce the model to over-fit to the synthetic domain, thus impairing the performance on realistic data. Thus, we set $\lambda$=0.8, after which, as shown in Fig.~\ref{fig:sensitivity} (b), it can be observed that the performance improves as $\textit{n}_s$ becomes larger and remains nearly the same when $\textit{n}_s$ $\geq$500k since it introduces enough objects from the novel categories for training. Hence, we set $\textit{n}_s$=500k to balance performance and efficiency of training. Further, we conducted a detailed investigation on the impact of LLMs. The results can be found in the Appendix, where our method achieves similar performance across different LLMs and demonstrates great robustness.

\noindent\textbf{Qualitative analysis.} We provide the visualization comparison between FC-CLIP and FC-CLIP+DreamMask on ADE20K validation set in Fig.~\ref{fig:qualitative}. It can be observed that our FC-CLIP+DreamMask produces much more accurate segmentation masks than FC-CLIP in both indoor and outdoor scenarios, especially on the novel categories, \ie, the windowpane in the first image of Fig.~\ref{fig:qualitative}.

\vspace{-1mm}
\section{Limitation}
\vspace{-1mm}
DreamMask has achieved significant performance boost on novel classes by expanding the training vocabulary. However, even though it makes an attempt to select samples with more reasonable and realistic context, there still exists room for improvements due to the sometimes fragile layout-to-image diffusion models when faced with synthesizing samples with complex scene descriptions. We do note that there exist some works that utilize in-context learning of LLMs to perform layout planning, integrating object-interaction diffusion for highly faithful image synthesis. We will defer this to future work.
    

\section{Conclusion}
In this paper, we present a novel sample imagination strategy, namely DreamMask, for open-vocabulary panoptic segmentation. It integrates LLMs and diffusion models to enrich the training set with context-aware novel samples. A multi-stage filtering strategy is introduced to further mitigate the negative impact of domain shifts by effectively retaining high-quality data. In addition, a synthetic-real alignment strategy is designed, which utilizes online-updating prototypes to close the gaps between the representations of the real and synthetic objects. Extensive experiments demonstrate that it can be integrated with existing methods to significantly boost their performance on novel categories. DreamMask also demonstrates superiority over training with web-crawled data, opening up a new direction for utilizing LLMs and diffusion models in other open-vocabulary tasks.

\medskip

{\small
\bibliographystyle{ieee_fullname}
\bibliography{egbib}

\begin{thebibliography}{10}\itemsep=-1pt

\bibitem{antoniou2017data}
Antreas Antoniou, Amos Storkey, and Harrison Edwards.
\newblock Data augmentation generative adversarial networks.
\newblock {\em arXiv:1711.04340}, 2017.

\bibitem{azizi2023synthetic}
Shekoofeh Azizi, Simon Kornblith, Chitwan Saharia, Mohammad Norouzi, and David~J Fleet.
\newblock Synthetic data from diffusion models improves imagenet classification.
\newblock {\em TMLR}, 2023.

\bibitem{brown2020language}
Tom Brown, Benjamin Mann, Nick Ryder, Melanie Subbiah, Jared~D Kaplan, Prafulla Dhariwal, Arvind Neelakantan, Pranav Shyam, Girish Sastry, Amanda Askell, et~al.
\newblock Language models are few-shot learners.
\newblock In {\em NeurIPS}, 2020.

\bibitem{bucher2019zero}
Maxime Bucher, Tuan-Hung Vu, Matthieu Cord, and Patrick P{\'e}rez.
\newblock Zero-shot semantic segmentation.
\newblock In {\em NeurIPS}, 2019.

\bibitem{caesar2018coco}
Holger Caesar, Jasper Uijlings, and Vittorio Ferrari.
\newblock Coco-stuff: Thing and stuff classes in context.
\newblock In {\em CVPR}, 2018.

\bibitem{chen2023open}
Xi Chen, Shuang Li, Ser-Nam Lim, Antonio Torralba, and Hengshuang Zhao.
\newblock Open-vocabulary panoptic segmentation with embedding modulation.
\newblock In {\em ICCV}, 2023.

\bibitem{chen2020uniter}
Yen-Chun Chen, Linjie Li, Licheng Yu, Ahmed El~Kholy, Faisal Ahmed, Zhe Gan, Yu Cheng, and Jingjing Liu.
\newblock Uniter: Universal image-text representation learning.
\newblock In {\em ECCV}, 2020.

\bibitem{cheng2020panoptic}
Bowen Cheng, Maxwell~D Collins, Yukun Zhu, Ting Liu, Thomas~S Huang, Hartwig Adam, and Liang-Chieh Chen.
\newblock Panoptic-deeplab: A simple, strong, and fast baseline for bottom-up panoptic segmentation.
\newblock In {\em CVPR}, 2020.

\bibitem{cho2023cat}
Seokju Cho, Heeseong Shin, Sunghwan Hong, Seungjun An, Seungjun Lee, Anurag Arnab, Paul~Hongsuck Seo, and Seungryong Kim.
\newblock Cat-seg: Cost aggregation for open-vocabulary semantic segmentation.
\newblock {\em arXiv:2303.11797}, 2023.

\bibitem{dao2023class}
Son~D Dao, Hengcan Shi, Dinh Phung, and Jianfei Cai.
\newblock Class enhancement losses with pseudo labels for open-vocabulary semantic segmentation.
\newblock {\em IEEE Transactions on Multimedia}, 2023.

\bibitem{devlin2019bert}
Jacob Devlin, Ming-Wei Chang, Kenton Lee, and Kristina Toutanova.
\newblock {BERT}: Pre-training of deep bidirectional transformers for language understanding.
\newblock In {\em NAACL}, 2019.

\bibitem{dhariwal2021diffusion}
Prafulla Dhariwal and Alexander Nichol.
\newblock Diffusion models beat gans on image synthesis.
\newblock In {\em NeurIPS}, 2021.

\bibitem{ding2022open}
Zheng Ding, Jieke Wang, and Zhuowen Tu.
\newblock Open-vocabulary universal image segmentation with maskclip.
\newblock In {\em ICML}, 2023.

\bibitem{dockhorn2022score}
Tim Dockhorn, Arash Vahdat, and Karsten Kreis.
\newblock Score-based generative modeling with critically-damped langevin diffusion.
\newblock In {\em ICLR}, 2022.

\bibitem{everingham2010pascal}
Mark Everingham, Luc Van~Gool, Christopher~KI Williams, John Winn, and Andrew Zisserman.
\newblock The pascal visual object classes (voc) challenge.
\newblock {\em IJCV}, 2010.

\bibitem{feng2023layoutgpt}
Weixi Feng, Wanrong Zhu, Tsu-jui Fu, Varun Jampani, Arjun Akula, Xuehai He, Sugato Basu, Xin~Eric Wang, and William~Yang Wang.
\newblock Layoutgpt: Compositional visual planning and generation with large language models.
\newblock In {\em NeurIPS}, 2023.

\bibitem{ghiasi2022scaling}
Golnaz Ghiasi, Xiuye Gu, Yin Cui, and Tsung-Yi Lin.
\newblock Scaling open-vocabulary image segmentation with image-level labels.
\newblock In {\em ECCV}, 2022.

\bibitem{Han_2023_ICCV}
Cong Han, Yujie Zhong, Dengjie Li, Kai Han, and Lin Ma.
\newblock Open-vocabulary semantic segmentation with decoupled one-pass network.
\newblock In {\em ICCV}, 2023.

\bibitem{he2023synthetic}
Ruifei He, Shuyang Sun, Xin Yu, Chuhui Xue, Wenqing Zhang, Philip Torr, Song Bai, and Xiaojuan Qi.
\newblock Is synthetic data from generative models ready for image recognition?
\newblock In {\em ICLR}, 2023.

\bibitem{he2023primitive}
Shuting He, Henghui Ding, and Wei Jiang.
\newblock Primitive generation and semantic-related alignment for universal zero-shot segmentation.
\newblock In {\em CVPR}, 2023.

\bibitem{ho2020denoising}
Jonathan Ho, Ajay Jain, and Pieter Abbeel.
\newblock Denoising diffusion probabilistic models.
\newblock In {\em NeurIPS}, 2020.

\bibitem{jiao2023learning}
Siyu Jiao, Yunchao Wei, Yaowei Wang, Yao Zhao, and Humphrey Shi.
\newblock Learning mask-aware clip representations for zero-shot segmentation.
\newblock In {\em NeurIPS}, 2023.

\bibitem{jiao2024collaborative}
Siyu Jiao, Hongguang Zhu, Jiannan Huang, Yao Zhao, Yunchao Wei, and Shi Humphrey.
\newblock Collaborative vision-text representation optimizing for open-vocabulary segmentation.
\newblock In {\em ECCV}, 2024.

\bibitem{karazija2023diffusion}
Laurynas Karazija, Iro Laina, Andrea Vedaldi, and Christian Rupprecht.
\newblock Diffusion models for zero-shot open-vocabulary segmentation.
\newblock {\em arXiv:2306.09316}, 2023.

\bibitem{devlin2018bert}
Jacob Devlin Ming-Wei~Chang Kenton and Lee~Kristina Toutanova.
\newblock Bert: Pre-training of deep bidirectional transformers for language understanding.
\newblock In {\em NAACL}, 2019.

\bibitem{kirillov2019panoptic}
Alexander Kirillov, Ross Girshick, Kaiming He, and Piotr Doll{\'a}r.
\newblock Panoptic feature pyramid networks.
\newblock In {\em CVPR}, 2019.

\bibitem{kirillov2023segment}
Alexander Kirillov, Eric Mintun, Nikhila Ravi, Hanzi Mao, Chloe Rolland, Laura Gustafson, Tete Xiao, Spencer Whitehead, Alexander~C Berg, Wan-Yen Lo, et~al.
\newblock Segment anything.
\newblock In {\em ICCV}, 2023.

\bibitem{lan2024proxyclip}
Mengcheng Lan, Chaofeng Chen, Yiping Ke, Xinjiang Wang, Litong Feng, and Wayne Zhang.
\newblock Proxyclip: Proxy attention improves clip for open-vocabulary segmentation.
\newblock In {\em ECCV}, 2024.

\bibitem{lei2021less}
Jie Lei, Linjie Li, Luowei Zhou, Zhe Gan, Tamara~L Berg, Mohit Bansal, and Jingjing Liu.
\newblock Less is more: Clipbert for video-and-language learning via sparse sampling.
\newblock In {\em CVPR}, 2021.

\bibitem{li2022language}
Boyi Li, Kilian~Q Weinberger, Serge Belongie, Vladlen Koltun, and Ren{\'e} Ranftl.
\newblock Language-driven semantic segmentation.
\newblock In {\em ICLR}, 2022.

\bibitem{li2020unicoder}
Gen Li, Nan Duan, Yuejian Fang, Ming Gong, and Daxin Jiang.
\newblock Unicoder-vl: A universal encoder for vision and language by cross-modal pre-training.
\newblock In {\em AAAI}, 2020.

\bibitem{liang2022open}
Feng Liang, Bichen Wu, Xiaoliang Dai, Kunpeng Li, Yinan Zhao, Hang Zhang, Peizhao Zhang, Peter Vajda, and Diana Marculescu.
\newblock Open-vocabulary semantic segmentation with mask-adapted clip.
\newblock In {\em CVPR}, 2023.

\bibitem{lin2014microsoft}
Tsung-Yi Lin, Michael Maire, Serge Belongie, James Hays, Pietro Perona, Deva Ramanan, Piotr Doll{\'a}r, and C~Lawrence Zitnick.
\newblock Microsoft coco: Common objects in context.
\newblock In {\em ECCV}, 2014.

\bibitem{liu2019end}
Huanyu Liu, Chao Peng, Changqian Yu, Jingbo Wang, Xu Liu, Gang Yu, and Wei Jiang.
\newblock An end-to-end network for panoptic segmentation.
\newblock In {\em CVPR}, 2019.

\bibitem{liu2021swin}
Ze Liu, Yutong Lin, Yue Cao, Han Hu, Yixuan Wei, Zheng Zhang, Stephen Lin, and Baining Guo.
\newblock Swin transformer: Hierarchical vision transformer using shifted windows.
\newblock In {\em ICCV}, 2021.

\bibitem{qin2023freeseg}
Jie Qin, Jie Wu, Pengxiang Yan, Ming Li, Ren Yuxi, Xuefeng Xiao, Yitong Wang, Rui Wang, Shilei Wen, Xin Pan, et~al.
\newblock Freeseg: Unified, universal and open-vocabulary image segmentation.
\newblock In {\em CVPR}, 2023.

\bibitem{qu2023layoutllm}
Leigang Qu, Shengqiong Wu, Hao Fei, Liqiang Nie, and Tat-Seng Chua.
\newblock Layoutllm-t2i: Eliciting layout guidance from llm for text-to-image generation.
\newblock In {\em ACM MM}, 2023.

\bibitem{radford2021learning}
Alec Radford, Jong~Wook Kim, Chris Hallacy, Aditya Ramesh, Gabriel Goh, Sandhini Agarwal, Girish Sastry, Amanda Askell, Pamela Mishkin, Jack Clark, et~al.
\newblock Learning transferable visual models from natural language supervision.
\newblock In {\em ICML}, 2021.

\bibitem{rombach2022high}
Robin Rombach, Andreas Blattmann, Dominik Lorenz, Patrick Esser, and Bj{\"o}rn Ommer.
\newblock High-resolution image synthesis with latent diffusion models.
\newblock In {\em CVPR}, 2022.

\bibitem{sariyildiz2023fake}
Mert~B{\"u}lent Sar{\i}y{\i}ld{\i}z, Karteek Alahari, Diane Larlus, and Yannis Kalantidis.
\newblock Fake it till you make it: Learning transferable representations from synthetic imagenet clones.
\newblock In {\em CVPR}, 2023.

\bibitem{sharma2018conceptual}
Piyush Sharma, Nan Ding, Sebastian Goodman, and Radu Soricut.
\newblock Conceptual captions: A cleaned, hypernymed, image alt-text dataset for automatic image captioning.
\newblock In {\em ACL}, 2018.

\bibitem{souly2017semi}
Nasim Souly, Concetto Spampinato, and Mubarak Shah.
\newblock Semi supervised semantic segmentation using generative adversarial network.
\newblock In {\em ICCV}, 2017.

\bibitem{thomee2016yfcc100m}
Bart Thomee, David~A Shamma, Gerald Friedland, Benjamin Elizalde, Karl Ni, Douglas Poland, Damian Borth, and Li-Jia Li.
\newblock Yfcc100m: The new data in multimedia research.
\newblock {\em Communications of the ACM}, 2016.

\bibitem{trabucco2023effective}
Brandon Trabucco, Kyle Doherty, Max Gurinas, and Ruslan Salakhutdinov.
\newblock Effective data augmentation with diffusion models.
\newblock {\em arXiv:2302.07944}, 2023.

\bibitem{wang2023hierarchical}
Xudong Wang, Shufan Li, Konstantinos Kallidromitis, Yusuke Kato, Kazuki Kozuka, and Trevor Darrell.
\newblock Hierarchical open-vocabulary universal image segmentation.
\newblock In {\em NeurIPS}, 2023.

\bibitem{wang2024open}
Zhaoqing Wang, Xiaobo Xia, Ziye Chen, Xiao He, Yandong Guo, Mingming Gong, and Tongliang Liu.
\newblock Open-vocabulary segmentation with unpaired mask-text supervision.
\newblock {\em arXiv:2402.08960}, 2024.

\bibitem{wu2023clipself}
Size Wu, Wenwei Zhang, Lumin Xu, Sheng Jin, Xiangtai Li, Wentao Liu, and Chen~Change Loy.
\newblock Clipself: Vision transformer distills itself for open-vocabulary dense prediction.
\newblock {\em arXiv:2310.01403}, 2023.

\bibitem{xie2023mosaicfusion}
Jiahao Xie, Wei Li, Xiangtai Li, Ziwei Liu, Yew~Soon Ong, and Chen~Change Loy.
\newblock Mosaicfusion: Diffusion models as data augmenters for large vocabulary instance segmentation.
\newblock {\em arXiv:2309.13042}, 2023.

\bibitem{xu2022groupvit}
Jiarui Xu, Shalini De~Mello, Sifei Liu, Wonmin Byeon, Thomas Breuel, Jan Kautz, and Xiaolong Wang.
\newblock Groupvit: Semantic segmentation emerges from text supervision.
\newblock In {\em CVPR}, 2022.

\bibitem{xu2023open}
Jiarui Xu, Sifei Liu, Arash Vahdat, Wonmin Byeon, Xiaolong Wang, and Shalini De~Mello.
\newblock Open-vocabulary panoptic segmentation with text-to-image diffusion models.
\newblock In {\em CVPR}, 2023.

\bibitem{xu2023side}
Mengde Xu, Zheng Zhang, Fangyun Wei, Han Hu, and Xiang Bai.
\newblock Side adapter network for open-vocabulary semantic segmentation.
\newblock In {\em CVPR}, 2023.

\bibitem{yang2023freemask}
Lihe Yang, Xiaogang Xu, Bingyi Kang, Yinghuan Shi, and Hengshuang Zhao.
\newblock Freemask: Synthetic images with dense annotations make stronger segmentation models.
\newblock In {\em NeurIPS}, 2023.

\bibitem{yu2023fcclip}
Qihang Yu, Ju He, Xueqing Deng, Xiaohui Shen, and Liang-Chieh Chen.
\newblock Convolutions die hard: Open-vocabulary segmentation with single frozen convolutional clip.
\newblock In {\em NeurIPS}, 2023.

\bibitem{zhou2017scene}
Bolei Zhou, Hang Zhao, Xavier Puig, Sanja Fidler, Adela Barriuso, and Antonio Torralba.
\newblock Scene parsing through ade20k dataset.
\newblock In {\em CVPR}, 2017.

\end{thebibliography}
}

\end{document}